# Detecting Fake Points of Interest from Location Data

Syed Raza Bashir
Department of Computer Science
Ryerson University
Toronto, Canada
syedraza.bashir@ryerson.ca

Vojislav Misic
Department of Computer Science
Ryerson University
Toronto, Canada
vmisic@ryerson.ca

*Abstract*— The pervasiveness of GPS-enabled mobile devices and the widespread use of location-based services have resulted in the generation of massive amounts of geo-tagged data. In recent times, the data analysis now has access to more sources, including reviews, news, and images, which also raises questions about the reliability of Point-of-Interest (POI) data sources. While previous research attempted to detect fake POI data through various security mechanisms, the current work attempts to capture the fake POI data in a much simpler way. The proposed work is focused on supervised learning methods and their capability to find hidden patterns in location-based data. The ground truth labels are obtained through real-world data, and the fake data is generated using an API, so we get a dataset with both the real and fake labels on the location data. The objective is to predict the truth about a POI using the Multi-Layer Perceptron (MLP) method. In the proposed work, MLP based on data classification technique is used to classify location data accurately. The proposed method is compared with traditional classification and robust and recent deep neural methods. The results show that the proposed method is better than the baseline methods.

## I. INTRODUCTION

Geolocation uses location technologies such as Global Positioning System (GPS) or IP addresses to identify and track the location of connected electronic devices [1], [2]. Geolocation is useful for tracking the movements of a mobile phone user. Today, every mobile device can save its exact GPS position in a few metres with high precision. This useful information is often used to prove or refute evidence in criminal cases. Furthermore, a variety of apps, including gaming and social networking and government-developed apps, assist in tracking infected COVID-19 users by interacting with a centralized server using their GPS location. The goal of tracking geolocations is to prevent dissemination of fake information or to protect sensitive locations [3]. Despite so many advancements, we find a few issues with the current geolocation technology in smart devices, which are:

1. These apps may not detect the real information that is saved in a suspect's or infected person's devices (e.g., smartphone) and thus allow the device to store fabricated false positions on remote locations (i.e., Google timeline). When an individual's phone transmits its location to centralized servers, the data is not accurate and is essentially useless.

2. These apps allow the users to change location to anywhere, plan a map trajectory with customized movement speeds and thus manipulate location-based apps such as games, social platforms, etc.

Geolocation can also be easily manipulated in the cloud and edge servers [4]. Usually, an application requests a geolocation value from a GPS device driver. In this process, there are many vulnerable points to forge the current geolocation of the devices. Related studies have proposed various levels of Trusted Computing Base (TCB) [5], which refers to the combination of security mechanisms within a computer system, including hardware, firmware, and software, responsible for enforcing a security policy. While TCB is particularly useful for establishing system-wide information and security policies, implementing it is a time-consuming and costly process. It necessitates the installation of numerous security-related files on a system. The system must then ensure that the file system tree does not contain any files that clearly violate system security. It also requires regular updates, additions, and deletions of trusted files.

In this work, we take a straightforward approach to the geolocation vulnerability issue. We propose to detect vulnerable as well as false geolocations before they become part of the geolocation-based lifecycle. Our work in this study is motivated by the observations as discussed in the following example.

According to a Statistica[1] report in April 2018, Google Maps alone has approximately 154.4 million monthly users in the United States. Waze, Apple Maps, Mapquest, Google Earth, Yahoo Maps are other popular choices. Despite so much investment and advanced Google Maps infrastructure, it has become apparent that Google Maps contains many false business contacts created by companies claiming to be close by [6]. According to the advertisers, research experts, existing and former Google employees, Google Maps is overwhelmed with millions of bogus addresses and false names triggered by Google queries [6]. If the issues mentioned above are not solved on time, there could be an exponential number of fake addresses on Google Maps. Google has responded [6] by removing thousands of such counterfeit addresses. For example, they remove the business listings from Google Maps[2]. But even though many of the bogus geolocations are eliminated, the damage they have already caused is irreversible.

---

[1] https://www.statista.com/statistics/865413/most-popular-us-mapping-apps-ranked-by-audience/

Furthermore, the tactics of the producers of fake geolocations change over time, which we can refer to as data drifts and concept drifts [7]. In concept drifts, the values of target variables change over time. For example, the geolocation that is labelled as true becomes fake. In data drifts, the statistical properties of the data change, which may affect the model's predictions, and their correlation with other variables. Fake geolocations may also be caused by system negligence, which is likewise not taken into account.

In this paper, we propose a fake location-aware detection method that can detect the fake locations which are either generated from the users' devices or stored in a centralized server. We propose a supervised learning and classification algorithm based on a feedforward artificial neural network using a Multi-Layer Perceptron (MLP) network. Our proposed method uses a feature combination that has the advantage of being computationally simple compared to previous methods while still being robust in fake location detection. The overall system accuracy of the method is improved by optimizing the number of feature vectors per sample, checking the sensitivity of the hyperparameters in MLP and the total training samples. We summarize our contributions as:

1. We crawled a dataset that consists of geolocations with known GPS-precise locations. We also added the fake geolocations in the dataset. The dataset provides the ground truth labels for the geolocation accuracy.
2. We build a deep neural network-based prediction model that can classify the geolocations into real or fake. The experimental results on the dataset show the accuracy of our model over the baseline methods for the fake geolocation classification problem.

The rest of the paper is organized as follows: Section II describes related work. Section III presents the proposed methodology and Section IV describes the experimental setup. Section V discusses the results and analysis and Section VI is about the limitations. Finally, Section VII concludes the paper.

## II. RELATED WORK

Points of Interest (POI) are being applied to address so many problems it's hard to keep track of them all. People and businesses use POIs on a daily basis for a variety of reasons. The POI data can be used in a variety of industries, but their true value lies in their accuracy. When it comes to making mission-critical decisions, POIs must be precise and accurate.

POI datasets[3] are multi-sourced to provide accurate location and company information for businesses, leisure, and geographic features in many countries and territories around the world. To ensure global consistency and ease of use, each dataset is cross-referenced to identify relationships and insights, and a hierarchical classification scheme is used to ensure global consistency and ease of use. Despite the usefulness of these datasets, data labelling to these POIs is a time-consuming process, limiting the applicability of these datasets in research.

Google[4] scans millions of contributions with automated detection systems and machine learning models. If the content is found to violate the platform's policies, it is automatically removed. Despite such a robust design, the generality of Google detection systems is limited by the availability of a large amount of data, despite their robust design. The effect of fake POI has already spread too far by the time such data becomes available.

The sources (humans, bots, agencies) that generate fake local listings use the information found in emails, social-networks, and other online platforms. Previous research [6], [8] has looked into how these sources obtained email addresses and account passwords through bulk registration to create fake geolocations. Among these approaches are obtaining VoIP phone numbers from telephony companies, obtaining mailing addresses from postal offices to use as re-shipping hubs, and scamming people into work-from-home scams and similar schemes to receive their emails and addresses.

In some works [5], the Trusted Computing Base (TCB) enforces system-wide information security regulations. TCB allows user access to the trusted communication path by installing and using the TCB functions. This approach requires installing the TCB protocol on the host.

Black Hat App Search Optimization (ASO)[5] is fraudulent reviews and fake accounts in peer-opinion platforms, such as app stores. The goal of Black Hat ASO is to improve the visibility of apps to get more downloads quickly. It is found that fraudulent posting activities are also connected to fake geolocations [8], [9]. These works [8], [9] provide insights into various aspects of fraudulent activities, such as automated queries to Google, links manipulation, sneaky redirects, to detect fake listing. However, such assumptions about the working procedures of fraudsters are only based on empirical analysis and lack an automated process. Manual processing to detect and identify such listings is a time-consuming task.

Some works [10] also provide an overview of the range of geolocation detection techniques, such as IP-based geolocation techniques, which could potentially estimate the location of a visiting user and perform geolocation cloaking attacks. In contrast, we implement a simple and automatic method that can detect fake geolocation in an automated fashion without too much prior knowledge about the system and software infrastructures. We present an algorithmic approach to detecting fake geolocations. Our model can perform detection even in the absence of manual analysis as in the previous works.

---

[2] How we fight fake business profiles on Google Maps (blog.google)
[3] https://www.precisely.com/product/precisely-points-of-interest/precisely-points-of-interest
[4] https://www.blog.google/products/maps/google-maps-101-how-contributed-content-makes-maps-helpful/
[5] https://thetool.io/2018/black-hat-aso

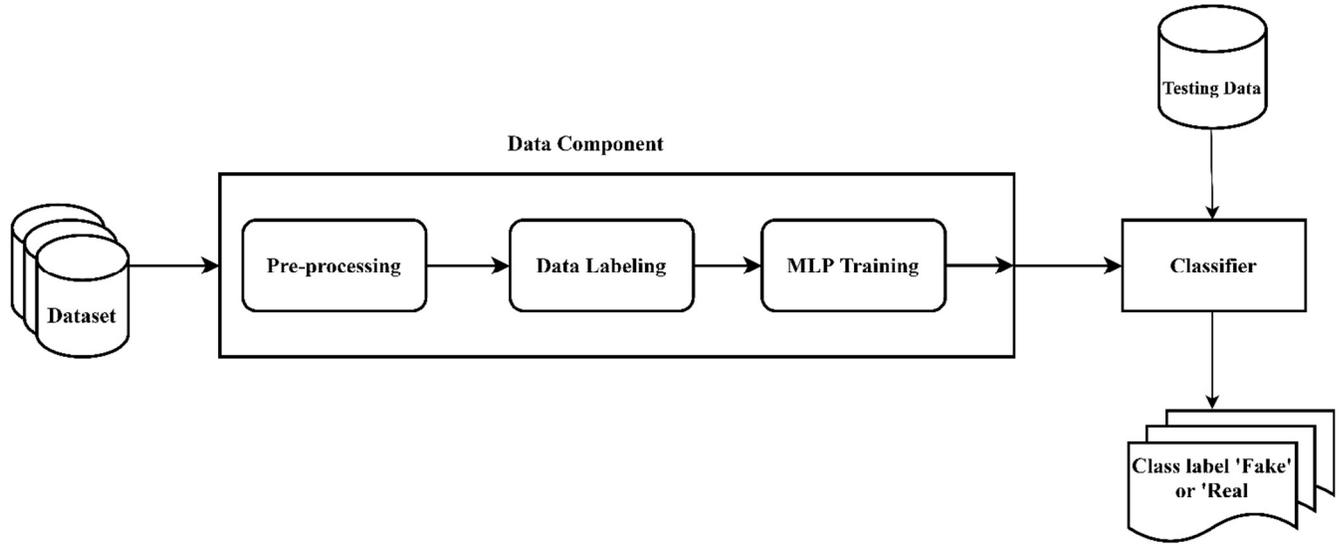

Fig. 1: General framework of Fake Point.

## III. METHODOLOGY

### A. Problem Definition

Given geolocation or POI, (In this paper, we use the term POI or geolocation interchangeably) the task of fake POI detection is to determine if a POI is fake or real. Formally, it can be defined as:

- Input: Geolocation or POI
- Output: One of two labels: "fake" or "real"

### B. Proposed Framework

We propose a deep learning-based detection system to detect POIs (as fake or real) from a given dataset. In this section, we briefly describe our proposed framework, that we name as Fake Point (FP), which is shown in Fig. 1. The framework consists of the following parts:

*Data component* takes as input the geolocation data. It has the following modules:

- Preprocessing module takes input from the data storage and preprocesses the data in a structured format.
- Data labelling module takes the structured data from the preprocessing module and performs the labelling task.
- MLP training module takes the labelled data and prepare it for model training.

*Classifier* takes the input from the training module and the testing data. The classifier, then fits the model on training data and evaluates on the test data. This module predicts if geolocation is fake or real, as its outcome.

Next, we dive into more details of the functionality and operation of each part of Fig. 1.

*Data component*: The component inputs the POI data from the data storage and prepares the data for the classification. We use a real-time POI data set of Regional Municipality of Peel[6] and added the synthetic information for the fake POI using Faker[7]. More details about the dataset are given in Section IV. It has preprocessing, data labeling and MLP training modules.

*Preprocessing*: This module processes the POI data that involves converting raw data into a structured format. The goal is to process and clean the data, detect missing values, handle imbalanced data, and detect duplicate records. The whole process transforms the data into the format that is ready to be input into the proposed deep neural network-based model. We perform a couple of steps in this regard: a) data cleaning by handling the missing data either by ignoring the tuple or filling the missing values manually where possible; b) cleaning noisy (erroneous) data; c) handling imbalanced data.

*Data Labelling:* The input to the data labelling module is the processed data, and output is the data that is labelled. We get the ground truth data from the Peel Region Data Center from the Government of Canada. The crawled data from this data centre is accurate and updated regularly, which is also available under an Open Data License. So, we consider the crawled data from the Peel Region Data Center as having true labels. The data using the Faker API is used to generate fake data. The usage of Faker to create fake data labels is also seem in the literature [11], so that portion of the data is considered to have 'fake' data labels.

*MLP Training*: This module prepares the data for training for the classifier model.

*Classifier:* The input to the classifier component is the labelled data with ground truth labels. This component consists of a deep neural network and is used to detect a POI as being 'true' or 'fake'. We use the MLP neural network in this module, discussed in Section III. This component takes the training part

---

[6] [Points of Interest | Data Portal - Region of Peel (peelregion.ca)](#)     [7] [Faker · PyPI](#)

in that is used to fit the model, and it also takes as input the test part that is used to evaluate the model. This component provides training to the neural network. This component's output is the POI labeled as 'true' or 'fake'.

*C. Proposed Methodology*

In this section, we discuss the proposed method and our classifier in detail. First, we discuss the preliminaries and then we go into the details of the model.

*1) Preliminaries*

A Perceptron is a supervised learning algorithm for binary classifiers. The MLP is a deep neural network that extends the perceptron to perform classification or regression, depending on its activation function [12]. We choose to work with MLP in this research because of its ability to solve the classification problem. While other neural networks can also be used for classification, in this study, we are more interested in the simplicity of the solutions while achieving a higher accuracy goal. Our intuition is to avoid excessive model complexity while maintaining some control and interpretability in the modelling process.

A MLP is a feedforward artificial neural network that produces outputs from a set of inputs. An MLP is described by several layers of input nodes connected as a directed graph between the input and output layers. MLP trains the network using backpropagation. Backpropagation is a standard algorithm for training feedforward neural networks that calculates the gradients of a loss function concerning all the weights in the network [13]. This paper uses the terms' activation function' and the 'loss function' a few times. The difference between the two functions is that the activation function activates the neuron required for the desired output and converts linear input to non-linear output. On the other hand, the loss function figures out the model's performance and finds how good the model can generalize; it computes the error for every training sample.

A MLP consists of an input layer, one or more hidden layers, and an output layer. The nodes (neurons) in the input layer correspond to the number of input variables in the processed data. When the data passes through the input layer, the values are weighted and processed onto the hidden layer. The output of the hidden layer can now be fed directly as input to another hidden layer or taken as output. The number of neurons in the output layer equals the number of outputs associated with each input.

The problem discussed in this work is to detect if a POI is real or fake. Next, we discuss the steps used in the classification algorithm in detail.

*2) Algorithm*

*Step 1*: Feed the input data to the input layer. The input layer consists of the neurons that receive inputs and pass them on to the other layers. The number of neurons in the input layer are equal to the attributes or features in the dataset. We represent our input by the explanatory variables (predictors) as $(x_1, x_2, \ldots, x_n) \in X$. The input is multiplied by the assigned weight values, and a bias value is added with each product, which is shown in Equation (1):

$$\alpha = \sum_{i=1}^{n} w_i x_i + b \quad (1)$$

Here $\alpha$ represents the weighted combination of inputs being aggregated, $w_i$ refers to the weight with the input variable $x_i$ and b represents the bias associated with weight. The weights are the strength or amplitude of a connection between two neurons, frequently initialized to small random values, such as 0.0 to 1.0.

*Step 2:* The input $\alpha$ is mapped to the output by an activation function *f*, as shown in Fig. 2.

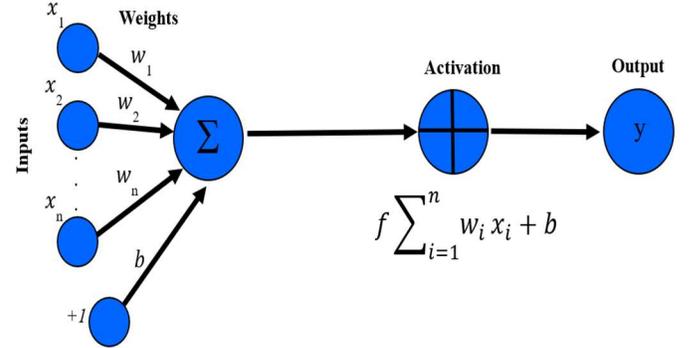

Fig. 2: Activation Function.

In Fig. 2, $(x_1, \ldots, x_n)$ is the signal vector that gets multiplied with the weights, $(w_1, w_2, \ldots, w_n)$. This is followed by accumulation (summation + addition of bias $b$). Finally, an activation function $f$ is applied to this sum. Note that the weights $(w_1, w_2, \ldots, w_n)$ and the bias $b$ transform the input signal linearly. Finally, the weighted sum obtained is turned into an output signal *y* by feeding the weighted sum into a non-linear activation function f (also called transfer function).

There are several activation functions for different use cases. In this work, we apply the Rectified Linear Unit (ReLU) activation function, which if the input is positive, outputs the input directly; otherwise, it outputs zero. ReLU is also easier to train and often achieves better performance [14] compared to other activation functions, such as sigmoid, logistic, tanh, erf, and similar. ReLu are also simple, faster to compute, and do not suffer from vanishing gradients, so we choose to work with this activation function in this work.

*Step 3:* We choose to work with cross-entropy [15], which is commonly used in classification tasks. Cross entropy loss function is an optimization function used for training classification models that classify the data by predicting the probability (value between 0 and 1) of whether the data belong to one class or another class. Cross-entropy loss is commonly used as the loss function for the models which has SoftMax output (a generalization of logistic regression to multiple dimensions) [15].

*Cross Entropy Classification:*

Considering $t^{(j)}$ and $O^{(j)}$ as the predicted (also known as target) output and actual output, respectively, for training example j; and y represent the output units and O the output layer, we define the loss function as shown in Equation (2):

$$L(W, B|j) = \sum_{y \in O} \left( ln(O_y^{(j)}) \cdot t_y^{(j)} + ln(1 - O_y^{(j)}) \cdot (1 - t_y^{(j)}) \right) \quad (2)$$

*Step 4:* In the next step, we minimize the loss function $L(W, B|j)$ during the model training. Usually, the Stochastic Gradient Descent (SGD) is used to minimize the loss function and optimize the model's hyperparameters [16]. Although stochastic gradient descent is quick and memory-efficient, it is difficult to parallelize without becoming slow. So, we use Hogwild [17], which is a lock-free parallelization scheme to address this issue of SGD. The Hogwild optimization algorithm is given below in Algorithm 1.

### Algorithm 1

1. Initialize global model parameters $W, B$
2. Distribute training data $\mathcal{T}$ across nodes (can be disjoint or replicated)
3. Iterate until convergence criterion reached:
    3.1. For nodes $n$ with training subset $\mathcal{T}_n$, do in parallel:
        a. Obtain copy of the global model parameters $W_n, B_n$
        b. Select active subset $\mathcal{T}_{na} \subset \mathcal{T}_n$ (user-given number of samples per iteration)
        c. Partition $\mathcal{T}_{na}$ into $\mathcal{T}_{nac}$ by cores $n_c$
        d. For cores $n_c$ on node $n$, do in parallel:
            i. Get training example $i \in \mathcal{T}_{nac}$
            ii. Update all weights $w_{jk} \in W_n$, biases $b_{jk} \in B_n$
            
            $$w_{jk} \coloneqq w_{jk} - \alpha \frac{\partial L(W, B|j)}{\partial w_{jk}}$$
            
            $$b_{jk} \coloneqq b_{jk} - \alpha \frac{\partial L(W, B|j)}{\partial b_{jk}}$$
    
    3.2. Set $W, B \coloneqq Avg_n\ W_n, Avg_n\ B_n$
    3.3. Optionally score the model on a (potentially sampled) train/validation scoring sets

In Algorithm 1, the weights and bias updates follow the asynchronous algorithm to adjust each node's parameters incrementally $W_n, B_n$ on example $i$. The $Avg$ notation represents the final averaging of these local parameters across all nodes to obtain the global model parameters and complete training.

## IV. EXPERIMENTAL SETUP

### A. Data Set

We scraped the geolocation dataset from Regional Municipality of Peel Open Data[8]. It contains around 1300 entries in different categories that range from institutional to public housing to early years centres. Some examples of the categories contained within the dataset are:

- Arts, museum, and cultural spaces
- Emergency responder stations: fire, police, and paramedics
- Institutional buildings: city/town halls, court houses, libraries
- Hospitals, medical centres, and walk-in clinics
- Housing: public housing, co-operative housing, shelters
- Food banks
- Long term care homes and retirement homes
- Post office
- Recreation centres and other municipal meeting places: arenas, pools, community centres, meeting halls
- Settlement services and other related services for immigrants and newcomers
- Shopping centres: plazas, big box centres, and malls
- Transportation: airports, major bus stations, and passenger rail stations

In the original dataset, there are 19 attributes, out of which we choose to work with some shown in Table I:

TABLE I. ATTRIBUTES USED.

| Attributes | Description |
|---|---|
| LM_ID | Landmark ID |
| X | Latitude |
| Y | Longitude |
| LM_NAME | Landmark Name |
| CATE | Category (type) of place |
| STR_ADD | Street Address |
| U | Unit number |
| MUN | Municipality |
| PR | Province |
| PC | Postal Code |
| PHONE | Phone number |
| WEBSITE | Website address |

We use the Faker[9] API to generate around 500 fake entries, on the top of original data. We classified the real POI as '1' and fake as '0'.

### B. Evaluation Metrics

Fake POI detection task is a binary decision problem, where the detection result is either fake or real geolocation. To assess the performance of our proposed model, we use the accuracy, precision, recall and F1-score as the evaluation metrics. The information about actual and predicted classifications is determined by the confusion matrix as shown in Table II. True Positive (TP) indicates that predicted fake samples are fake. False Positive (FP) indicates that predicted real samples are fake. False Negative (FN) indicates that predicted news samples are real. True Negatives (TN) indicates that predicted real samples are real.

TABLE II. CONFUSION MATRIX.

|  | Actual Fake | Actual Real |
|---|---|---|
| **Predicted Fake** | TP | FP |
| **Predicted Real** | FN | TN |

For the precision, recall, F1-score and accuracy, we perform the specific calculation as:

Accuracy is the overall performance of the model and can be defined as in Equation (3):

---

$$\text{Accuracy} = \frac{TP + TN}{TP + TN + FP + FN} \quad (3)$$

Recall is the coverage of actual positive samples and can be defined as in Equation (4):

$$\text{Recall} = \frac{TP}{TP + FN} \quad (4)$$

Precision indicates how accurate are the positive predictions in Equation (5):

$$\text{Precision} = \frac{TP}{TP + FP} \quad (5)$$

F1 score indicates the hybrid metric useful for unbalanced classes in Equation (6):

$$\text{F1 score} = \frac{2TP}{2TP + FP + FN} \quad (6)$$

We also test our model using Root Mean Square Error (RMSE) is a standard method of calculating a model's error in predicting quantitative data. It measures the difference between values predicted by a model and the actual values observed from the environment that is being modelled.

We divide the dataset into three different parts: training set (70%), validation set (15%) and testing set (15%). We use the $k$-fold cross-validation to select the model parameters. We take 5 $k$ folds. To handle the data imbalance problem, we use the Synthetic Minority Oversampling Technique (SMOTE) [18], where the synthetic samples are generated for the minority class. We have used the adaptive learning rate algorithm ADADELTA [19] that combines the benefits of learning rate and momentum training to avoid slow convergence.

*C. Hyperparameters*

The hyperparameters used in this model are in Table III.

TABLE III. HYPERPARAMETERS WITH VALUES.

| Hyperparameter | Value |
|---|---|
| Hidden layer size | 200 |
| Epochs | 10 |
| Seed | -1 |
| Adaptive learning rate | Enabled |
| Adaptive learning rate time decay factor | 0.99 |
| Epsilon for adaptive_rate | 1e-08 |
| Dropout ratio | 0.5 |
| L1 regularization | 1e-5 |
| L2 regularization | 1e-5 |
| Activation function | ReLu |
| Optimizer | Adadelta |
| Adaptive learning:$\rho$(rho) for Adadelta | 0.99 |
| Adaptive learning:$\epsilon$(epsilon), for Adadelta | 0.99 |

We tested different L1 and L2 regularization values on the RMSE and reported the result with the best performing values. The performance of the model with varying values of L1-regularization is shown below in Fig. 3. While using one value for regularization, the other is fixed. We conduct experiments in parallel with many values of the regularization terms. To explain the significance of each of these values and their importance in the sensitivity analysis, we are showing these results sequentially in Fig. 3 and Fig. 4.

The result in Fig. 3 shows the best performance of our model when the strength of L1 regularization is 1.0E-05, so we choose to work with this value of L1-regularization in our experiments.

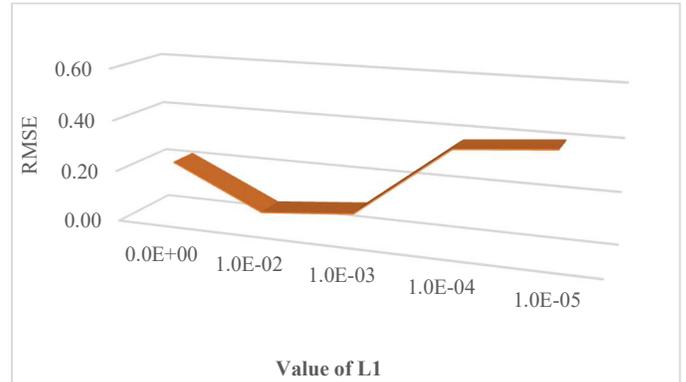

Fig. 3: RMSE as function of the value of L1.

We also test the performance of our model with l2-regularization and show the results in Fig. 4:

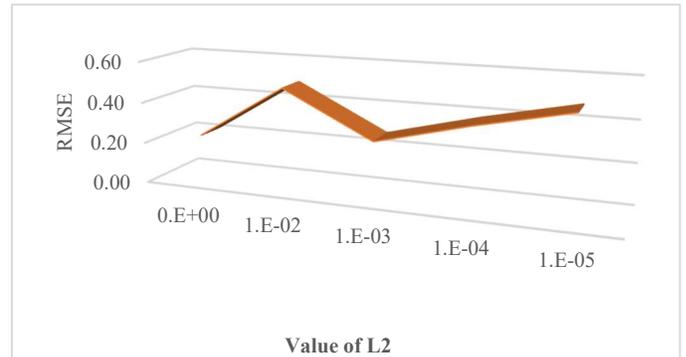

Fig. 4: RMSE as function of the value of L2.

We see the best performance of the model with L2-regularization value at 1.E-02, so we use this value in for our model. We also test various hyperparameters like dropout rate, adaptive learning rate and different number of epochs and report the best performing hyperparameters in Table III.

*D. Baselines*

We have used a mix of machine learning and deep neural network-based baselines. Our chosen baselines are listed below:

BERT (Bidirectional Encoder Representations from Transformers) [20]: BERT is a Google-developed transformer-based machine learning technique for pre-training in natural language processing (NLP). We use the cased and uncased versions of BERT.

Funnel Transformer [21]: Funnel Transformer is a bidirectional transformer model similar to BERT, but with a pooling operation after each block of layers, similar to how traditional convolutional neural networks (CNN) in computer vision work.

BART (Bidirectional and Auto-Regressive Transformer) [22]: BART is a Seq2Seq model that combines a Bidirectional Encoder (i.e. BERT) with an Autoregressive Decoder (i.e. GPT).

RoBERTa (Robustly Optimized BERT Pretraining Approach) [23]: RoBERTa model uses the original BERT model and improves BERT for training the model longer, with bigger batches, over more data.

Naive Bayes [24]: Naive Bayes is a probabilistic machine learning algorithm based on the Bayes algorithm and used for classification tasks.

SVM (Support Vector Machine) [25]: SVM is a supervised machine learning algorithm that can be used to solve classification and regression problems. We use the Linear Kernel SVM.

LDA (Linear Discriminant Analysis) [26]: LDA is a tool for data visualization, classification, and dimension reduction.

Logistic Regression [27]: Logistic Regression is a statistical analysis method for predicting a data value based on previous data set observations.

Decision Tree Classifier [28] : A simple representation for classifying examples is a Decision Tree. It's a type of Supervised Machine Learning in which data is continuously split according to a parameter.

Extra Trees Classifier [29]: Extra Trees Classifier it is an ensemble learning method that uses decision trees as its foundation.

Random Forest Classifier [30]: Random Forest Classifier is an ensemble tree-based learning algorithm. It is a set of decision trees derived from a subset of the training set chosen at random.

AdaBoost Classifier (Adaptive Boosting) [31]: AdaBoost is an iterative ensemble method. The AdaBoost classifier creates a strong classifier by combining several low-performing classifiers, resulting in a high-accuracy classifier.

Gradient Boosting Classifier [32]: Gradient Boosting Classifier is a set of machine learning algorithms that combine a number of weak learning models to produce a powerful predictive model.

Light GBM (Light Gradient Boosting Machine) [33]: Light GBM is an open-source library that implements the gradient boosting algorithm efficiently and effectively.

QDA (Quadratic Discriminant Analysis) [34]: QDA is a well known supervised classification methods in statistical and probabilistic learning.

k-NN Classifier (k-Nearest Neighbors) [35]: k-NN Classifier is a machine learning algorithm that is very simple, easy to understand, and versatile. It is based on the feature similarity approach.

We optimize the hyperparameters settings for each baseline and report the results using optimal hyperparameters.

## V. RESULTS AND ANALYSIS

### A. Model Performance

We show the learning curve for training loss and validation loss during the model training as shown in Fig. 5.

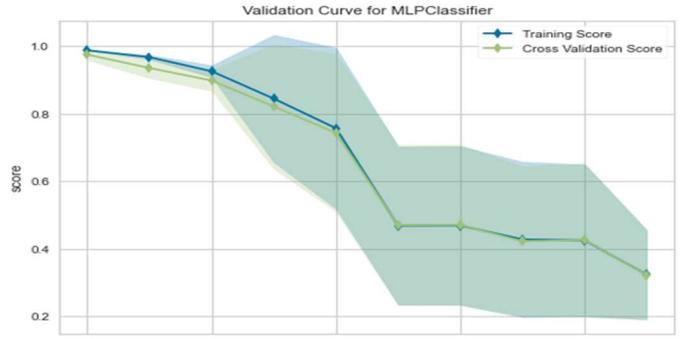

Fig. 5: Learning Curve for Training Loss and Validation Loss.

As shown in Fig. 5, the validation loss is quite close to the training loss. This shows a good fit for model training, defined as a training and validation loss that gradually decreases to the point of stability with a small gap between the two final loss values [36]. Overall, the result in this experiment explains how well it matches a set of observations. In most cases, such goodness of fit indicator summarizes the difference between observed and model-predicted values.

We also show the model performance on the test set and the confusion matrix is shown below in Table IV.

TABLE IV. CONFUSION MATRIX ON TEST SET.

|  | Actual Fake | Actual Real |
|---|---|---|
| **Predicted Fake** | 365 | 23 |
| **Predicted Real** | 11 | 168 |

The results in Table IV shows that our model has 94% precision, 97% recall and 95% F1-score. The model's accuracy is also 94%, which shows that 94% of the predictions are correct. The precision of 94% means that we have fewer false positives (POI is real but predicted as fake) and we can predict a large number of true positives (POI is fake and predicted as fake). The recall of 97% means we have many true positives compared to the false negatives. Generally, a false negative (POI is fake but predicted as real) is a worse error than a false positive in the POI detection problem. Overall, we get fewer false negatives than false positives (which are also fewer), that's why we get quite a high F1-score.

### B. Overall Model Performance

The results and analysis of the proposed model against the baselines are shown in Table V.

Overall, we can see that our proposed model has the highest accuracy (95%), precision (94%), recall (95%), F1-score (92%) and the average precision (69%) among all the baselines. The superiority of our model in detecting the fake POI is attributed to its design that is built upon a simple but carefully optimized neural network model.

The Naive Bayes, SVM - Linear Kernel, Linear Discriminant Analysis, Logistic Regression are machine learning models that perform the second-best in terms of accuracy, precision, and recall F1-score. This group of baselines usually shows better performance when there is a clear margin of separation between classes, which we can see in our data.

TABLE V. PROPOSED METHOD AND BASELINES RESULTS.

| Baselines | Accuracy | Recall | Precision | F1 |
|---|---|---|---|---|
| **Fake Detection Model (FDM)** | **0.94** | **0.97** | **0.94** | **0.95** |
| BERT (uncased) | 0.78 | 0.75 | 0.64 | 0.68 |
| BERT (cased) | 0.65 | 0.61 | 0.53 | 0.57 |
| Funnel Transformer | 0.50 | 0.68 | 0.52 | 0.59 |
| BART | 0.45 | 0.53 | 0.45 | 0.49 |
| RoBERTa | 0.45 | 0.51 | 0.46 | 0.48 |
| Naive Bayes | 0.83 | 0.73 | 0.69 | 0.71 |
| SVM - Linear Kernel | 0.82 | 0.82 | 0.85 | 0.84 |
| LDA | 0.81 | 0.81 | 0.85 | 0.83 |
| Logistic Regression | 0.80 | 0.80 | 0.84 | 0.82 |
| Extra Trees Classifier | 0.79 | 0.79 | 0.85 | 0.82 |
| Decision Tree Classifier | 0.79 | 0.78 | 0.85 | 0.81 |
| Random Forest Classifier | 0.79 | 0.78 | 0.85 | 0.81 |
| AdaBoost Classifier | 0.78 | 0.78 | 0.84 | 0.81 |
| Gradient Boosting Classifier | 0.77 | 0.76 | 0.85 | 0.80 |
| Light GBM | 0.76 | 0.79 | 0.81 | 0.80 |
| QDA | 0.75 | 0.77 | 0.80 | 0.79 |
| k-NN Classifier | 0.72 | 0.70 | 0.84 | 0.76 |

The Extra Trees, Decision Tree, Random Forest, AdaBoost, Gradient Boosting, Light Gradient Boosting Machine, Quadratic Discriminant Analysis and K Neighbors classifiers are the third best performing models, which are also machine learning models. The random forest and decision tree-based models build multiple decision trees and merges them together to get a more accurate and stable prediction. This exactly shows why some of these tree-based models perform better. The gradient boosting based algorithms also perform well when there is not much noise in the data (as in our dataset). Our data is already clean and there are not any outliers, so these models demonstrate their best.

The BERT (uncased), BERT (cased), Funnel Transformer, BART and RoBERTa have demonstrated modest performance among the neural baselines. This is probably because these models are pre-trained on huge corpora that lack location-aware or POI data. Typically, these transformer-based models perform well in tasks where there is lot of textual data and when the test data matches with the vocabulary of the pre-trained data. In our setup, the data is unique (Peel region location data), much of which is not possible to get from the pre-trained checkpoints like Wikipedia corpus (as in BERT) or similar datasets.

### C. Ablation Study

In this work, we try to show the feature importance for different attributes through an ablation study. The significance of each feature is a score based on how much a specific feature has improved a model's accuracy. We perform an ablation study on our model by removing different features from the data one at a time or in a group to see how it affects the model's performance.

The default model name is FDM, while model variants during the ablation study are labelled using the following convention. When we remove a part or feature of data, we describe it with the feature name, followed by a minus sign. For example, FDM(LM-) means FDM without LM_Name (see Table I for attribute names). We have tested the model by removing select features as well as select pairs of features, and the results are shown in Table VI:

TABLE VI. ABLATION STUDY ON MODEL VARIANTS.

| Model Variant | RMSE |
|---|---|
| FDM | 0.012 |
| FDM (LM_ID -) | 0.104 |
| FDM (X -, Y -) | 0.191 |
| FDM (LM_NAME -) | 0.176 |
| FDM (CATE -) | 0.144 |
| FDM (STR_ADD -) | 0.143 |
| FDM (MUN -) | 0.176 |
| FDM (PR -) | 0.185 |
| FDM (PC-) | 0.190 |
| FDM (STR_ADD -, U -) | 0.115 |
| FDM (MUN -, PC -) | 0.312 |
| FDM (MUN -, PR -, PC -) | 0.416 |
| FDM (STR_ADD -, U -, MUN -) | 0.465 |
| FDM (STR_ADD -, U -, MUN -, PR -) | 0.612 |
| FDM (STR_ADD -, U -, MUN -, PR -, PC -) | 0.714 |

The results in Table VI show that the default FDM model has the lowest RMSE score, which shows that the model performs best when we consider all the data features. The model's performance is degraded when we remove the x and y-axis that indicates the location axis. Model performance is also more negatively impacted when we remove the postal code. This is understandable as the fake POI are primarily generated using fake postal codes.

As shown in Table VI, the model's performance is most impacted when we remove many features from the data. This is understandable because when we give fewer features to the model, the model will not have enough information to generalize and to make accurate predictions.

The results also show when we remove the landmark name (LM), the model performance is not much impacted. This is probably because the location name is trivial. The location axis (x, y) is a more important location indicator and can be used to determine the truth about a location.

We also see that when we individually remove the street address, municipal and province information, the model performance is not much effected. This is because when we remove these pieces of data individually, the other related features have enough information to predict the location truth. However, suppose we remove a combination of these pieces of information, such as street address, province, municipality. In that case, the model's performance is impacted negatively (as seen in FDM (STR_ADD -, U -, MUN -, PR -, PC -)) . This is because removing too much of this information altogether tends to weaken the model's predictive power.

The dependencies of the attributes on each other also impact the model performance. For example, street address, municipality, the unit number are related attributes that depend on each other to predict an outcome. These results probably indicate that when two features are combined, they are more significant in explaining relationships in the data than the same two attributes separately.

Overall, the result suggests we should consider more location-related features to detect the truth about a location. That is why the default FDM model shows the best performance.

## VI. LIMITATIONS

This work shows an attempt in finding a suitable classification system for POI detection. There are some limitations of data and methods that are important to note here.

First, we use regional data based on the criteria that we want to model this problem on real-world data. While this solves the problem, the dataset does not represent the fake POI detection at a vast scale. It is critical to building diverse and challenging datasets to inform better detectors against all types of fake POIs. We recommend expanding our dataset approach and developing a benchmark representing the veracity of location-based content in various applications.

Second, this detection method is limited by the features available at the time. This method does not take into account scenarios in which there are user reviews, comments, or concerns about the location's veracity. Future work will include analyzing textual content from various forums on real-time events, keywords/hashtags, opinions/reviews, hyperlinks, and similar on location-based data.

## VII. CONCLUSION

In this paper, we study detecting fake geolocations for various Points of Interest (POI). We get the real-world data for ground truth labels, and we used the Faker API to generate synthetic fake data. Our model is based on MLP neural network, and we treat the problem of fake POI as binary classification. Through detailed experiments, we show the superiority of our model over several baseline models. Through ablation study, we show the importance of various features to be included in the model. In future, we plan to extend the model to include more real-world data, and we also plan to expand our model to other deep neural networks or towards an ensemble approach.